\documentclass[conference]{IEEEtran}
\IEEEoverridecommandlockouts
\usepackage{cite}
\usepackage{hyperref}
\usepackage{amsmath,amssymb,amsfonts}
\usepackage{algorithmic}
\usepackage{graphicx}
\usepackage{textcomp}
\usepackage{xcolor}
\usepackage{xurl}
\def\BibTeX{{\rm B\kern-.05em{\sc i\kern-.025em b}\kern-.08em
    T\kern-.1667em\lower.7ex\hbox{E}\kern-.125emX}}

\makeatletter

\def\ps@IEEEtitlepagestyle{%
    \def\@oddfoot{\mycopyrightnotice}%
    \def\@evenfoot{}%
}
\def\mycopyrightnotice{%
    {\footnotesize  979-8-3503-5901-5/23/\$31.00 \textcopyright2023 IEEE\hfill}
    \gdef\mycopyrightnotice{}
}

\makeatletter
\newcommand*\titleheader[1]{\gdef\@titleheader{#1}}
\AtBeginDocument{%
  \let\st@red@title\@title
  \def\@title{%
    \bgroup\normalfont\small\centering\@titleheader\par\egroup
    \vskip.5em\st@red@title}
}
\makeatother

\title{Explainable Contrastive and Cost-Sensitive Learning for Cervical Cancer Classification}

\titleheader{Accepted and presented in 26th International Conference on Computer and Information Technology (ICCIT 2023)}

\author{\IEEEauthorblockN{Ashfiqun Mustari, Rushmia Ahmed, Afsara Tasnim, Jakia Sultana Juthi and G. M. Shahariar}
\IEEEauthorblockA{Ahsanullah University of Science and Technology, Dhaka, Bangladesh\\
Email: \{asfiqunishaa, hridaa12, afsaratasnim131, jakiasultanajuthi04030, sshibli745\}@gmail.com}}

\begin{document}

\maketitle
\begin{abstract}
This paper proposes an efficient system for classifying cervical cancer cells using pre-trained convolutional neural networks (CNNs). We first fine-tune five pre-trained CNNs and minimize the overall cost of mis-classification by prioritizing accuracy for certain classes that have higher associated costs or importance. To further enhance the performance of the models, supervised contrastive learning is included to make the models more adept at capturing important features and patterns. Extensive experimentation are conducted to evaluate the proposed system on the SIPaKMeD dataset. The experimental results demonstrate the effectiveness of the developed system, achieving an accuracy of 97.29\%. To make our system more trustworthy, we have employed several explainable AI techniques to interpret how the models reached a specific decision. The implementation of the system can be found at - \url{https://github.com/isha-67/CervicalCancerStudy}.
\end{abstract}


\section{Introduction}
Cervical cancer, the world's third-most common type of cancer, is the leading cause of cancer-related deaths in women \cite{b1}. However, unlike other cancers, cervical cancer can be prevented. Many cytology-based screening programs can detect cervical cell abnormalities before they become cancer. One of the most popular screening tests for cervical cancer is the Papanicolaou test\cite{b2}, Pap test. However, a challenge in detecting cancerous cells from pap smear images is that it requires highly qualified pathologists to analyze the results, which is difficult to find, especially in developing countries. This is where computer-supported tools, such as deep learning, particularly CNNs, can be effective.

Convolutional Neural Networks(CNNs) automatically extract important features from input images, eliminating the need for manual feature extraction. Several studies \cite{b4,b5} have used convolutional neural networks (CNNs) to classify cervical cancer. But, a concern associated with these studies is that how the deep learning models make such decisions to classify the images cannot be explained. Some studies on medical images\cite{b6}\cite{b7} have utilized explainable AI techniques to explain how models reached a specific conclusion.

In our study, we developed a deep learning-based system for classifying cancer cell types using the SIPaKMeD Dataset\cite{b8}. We fine-tuned five different classifiers, refining their ability to classify cancer cell types. In order to make the classifiers more attuned to real-world classification errors, we included cost-sensitive learning. We also applied the concept of supervised contrastive learning\cite{b9} so that our models extract more discriminative and representative features and enhance the classification accuracy. In addition, we implemented explainable AI techniques to provide insights into our models' decision-making processes, aiming to build trust in automated medical image classification.

\section{Related Works}
Some existing studies have classified cervical cancer cells by utilizing convolutional neural networks. 
A. Ghoneim, G. Muhammad, and M. S. Hossain\cite{b10} developed a cervical cancer cell detection and classification system using pre-trained VGG16 and CaffeNet, utilizing the Herlev dataset. Pramanik et al.\cite{b5} developed an ensemble method that minimized the error between observations and ground truth, outperforming Inception V3, Inception ResNet V2, and MobileNetV2 models with an accuracy of 96.96\%. Manna, R. Kundu, D. Kaplun, A. Sinitca, and R. Sarkar\cite{manna2021fuzzy} introduced a classification model using ensemble methods, combining three CNN architectures: Inception v3, DenseNet-169, and Xception. The model achieved high accuracy rates of 98.55\% for a binary classification and, 95.43\% for a 5-class classification on the SIPaKMeD dataset and 99.23\% on the Mendeley LBC dataset. A. Tripathi, A. Arora, and A. Bhan\cite{b4} classified cancer cell growth stages using pre-trained models, with ResNet-152 achieving 94.89\% highest accuracy. On a different trajectory, Hsieh et al.\cite{b13} presented a different method for detecting bone metastases using contrastive learning. The study showed that contrastive learning enhances the accuracy of deep learning models. None of the studies considered the imbalance in data which Ravi\cite{b14} addressed in his studies by developing an attention-cost-sensitive deep learning-based feature fusion ensemble meta-classifier technique for skin cancer classification. He used cost weights to address data imbalance where his model outperformed other models with a 96\% accuracy rate. However, none of the studies provided any insight into interpreting how their best-performing models classified the images. Panwar et al.\cite{b6} leveraged the GradCAM-based color visualization approach to interpret radiology image detection on medical images using chest X-ray and CT scan images.
Notably, prior studies, despite showcasing promising performance utilizing CNNs, overlooked imbalanced data, interpretability, and the use of contrastive learning in the study of cervical cells detection. It highlights a specific gap in the current cervical cancer research that our study addresses, suggesting improved performance in cervical cancer cells classification.
\section{Methodology}
\begin{figure}[h]
    \centering
    \includegraphics[width=85mm,scale=1]{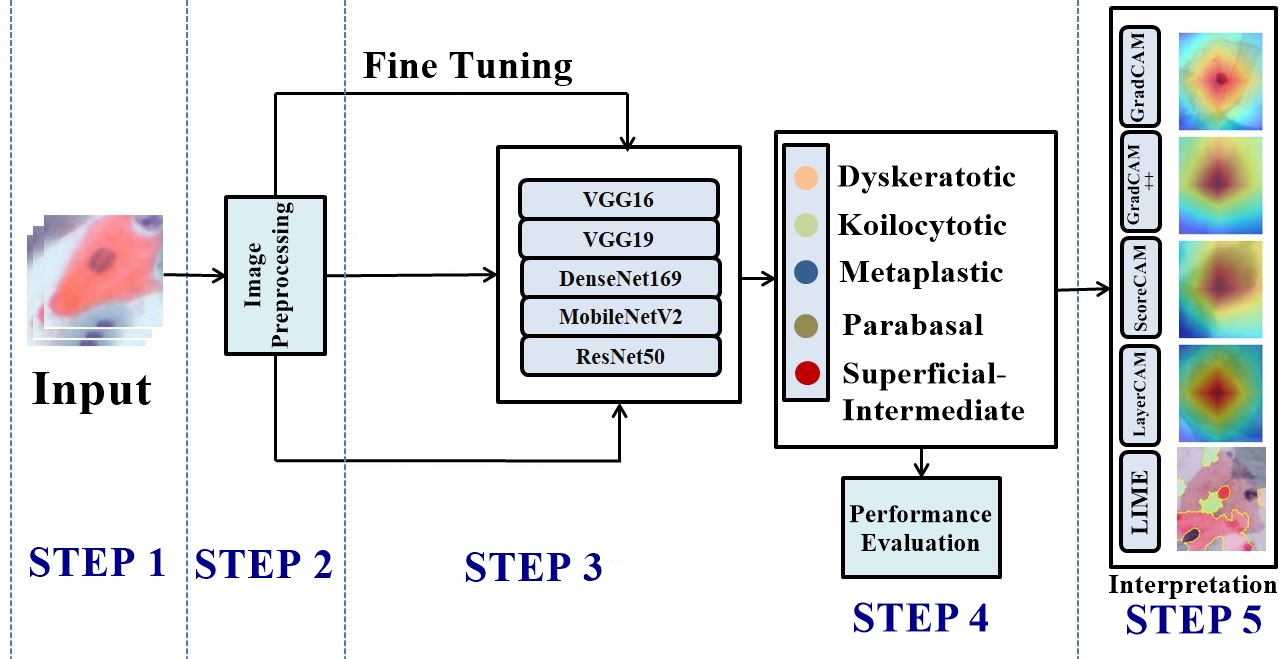}
    \caption{Proposed Methodology}
    \label{fig:Methodology}
\end{figure}

Figure \ref{fig:Methodology} presents our proposed methodology that comprises five distinct steps: input data, image preprocessing, model training, performance evaluation, and the interpretation of the model performance. We describe each of the steps below in detail.

\subsection{Input}The input to the system is the SIPaKMeD Dataset\cite{b8} consisting of pap smear images, where medical experts have manually defined the nucleus and the area of cytoplasm in each image and labeled the images. It includes 4049 precisely cropped photos of isolated cells and 966 images of Pap smear slide cluster cells. There are five different class labels: \textit{Superficial-Intermediate} (813 images), \textit{Parabasal} (727 images), \textit{Koilocytotic} (825 images), \textit{Metaplastic} (793 images) and \textit{Dyskeratotic} (813 images).
The dataset was split into three sets: train, test, and validation. The train set comprises 80\% of the total images. The remaining 20\% of the images were further divided into test (80\%) and validation (20\%) sets, resulting in 2589 train images, 648 validation images, and 812 test images. We augmented only the original training set, resulting in 18,123 images for a more diverse training dataset. Augmentations include affine transformations (rotation, translation, scaling, shearing, zooming, flipping, padding), noise injection, contrast adjustment, brightness modification, and pixel value changes.
\subsection{Image Preprocessing}For consistency and compatibility, all dataset images were standardized to a uniform resolution. The dataset contained images ranging from $62 \times 48$ pixels to $531 \times 553$ pixels. Downsizing large images can hinder feature learning, while up-scaled and zero-padded small images add complexity. We determined the optimal resolution through scatter and density plots. We calculated the dimensions of each image and created a 2D histogram to find the most common size, which was $98.5 \times 108.9$ pixels. We chose $110 \times 110$ pixels for uniformity, maintaining aspect ratios, facilitating further processing, and enhancing machine learning efficiency.

\subsection{Training}
We performed three types of training on five different pre-trained CNN models: standard fine-tuning, fine-tuning with cost-sensitive learning, and fine-tuning with supervised contrastive learning. This section provides a brief overview of the models, cost-sensitive and contrastive learning.

\subsubsection{Models}
For our experiment, we have utilized ResNet-50 \cite{b15}, MobileNetV2 \cite{b16}, DenseNet-169 \cite{b17}, VGG16 \cite{b18} and VGG19. We chose ResNet-50 as it is ideal for deep networks and has high accuracy due to skip connections. MobileNetV2 provides a balance between size and accuracy. DenseNet-169 promotes feature reuse, efficient parameter usage, and good accuracy. VGG16 and VGG19 are Simple, widely used architectures suitable for transfer learning and various image classification tasks.
During fine-tuning, we employed a layer-wise freezing strategy for different CNNs and achieved the maximum performance when we froze the first 86, 100, 249, 13, and 17 layers of ResNet-50, MobileNetV2, DenseNet-169, VGG16, and VGG19 respectively. This strategic freezing preserved pre-trained features, focusing updates on later layers to adapt the models for cervical cancer classification.

\subsubsection{Cost-Sensitive Learning} Cost-Sensitive learning is an area of learning where the costs associated with class-imbalanced data are taken into account \cite{b19,b20}. Depending on the application, there are different ways to create the cost matrix. One way is to put class weights according to the distribution of class labels. The default log loss function, as represented by Equation \ref{eq:eq1}, assigns equal weight to all classes in classification tasks, regardless of their distribution. This can introduce bias in cases of imbalanced data.
\begin{equation}
    LogLoss = 1/N \sum_{i=1}^{N} [-(y_i log(\Bar{y_i})+(1 - y_i) log(1-\Bar{y_i}))] \label{eq:eq1}
\end{equation}
To address this limitation, Equation \ref{eq:eq2} calculates the Weighted Log Loss by assigning proper weights to each class based on their distribution in unbalanced data.
\begin{align}
    \text{WeightedLogLoss} = \frac{1}{N} \sum_{i=1}^{N} &\left[ -w_0 (y_i \log(\Bar{y_i})) \right. \nonumber \\
    &+ \left. w_1 ((1 - y_i)\log(1-\Bar{y_i})) \right] \label{eq:eq2}
\end{align}
As the SIPaKMeD dataset is slightly imbalanced, we assign the costs of \textit{Dyskeratotic, Koilocytotic, Metaplastic, Parabasal, and Superficial-Intermediate} classes with values 0.996319018404908, 0.9842424242424243, 1.0213836477987421, 1.0278481012658227, and 0.9724550898203593, respectively, in all the experiments related to cost-sensitive learning.

\subsubsection{Supervised Contrastive Learning} 
\begin{figure}[t]
    \centering
    \includegraphics[width=80mm,scale=1]{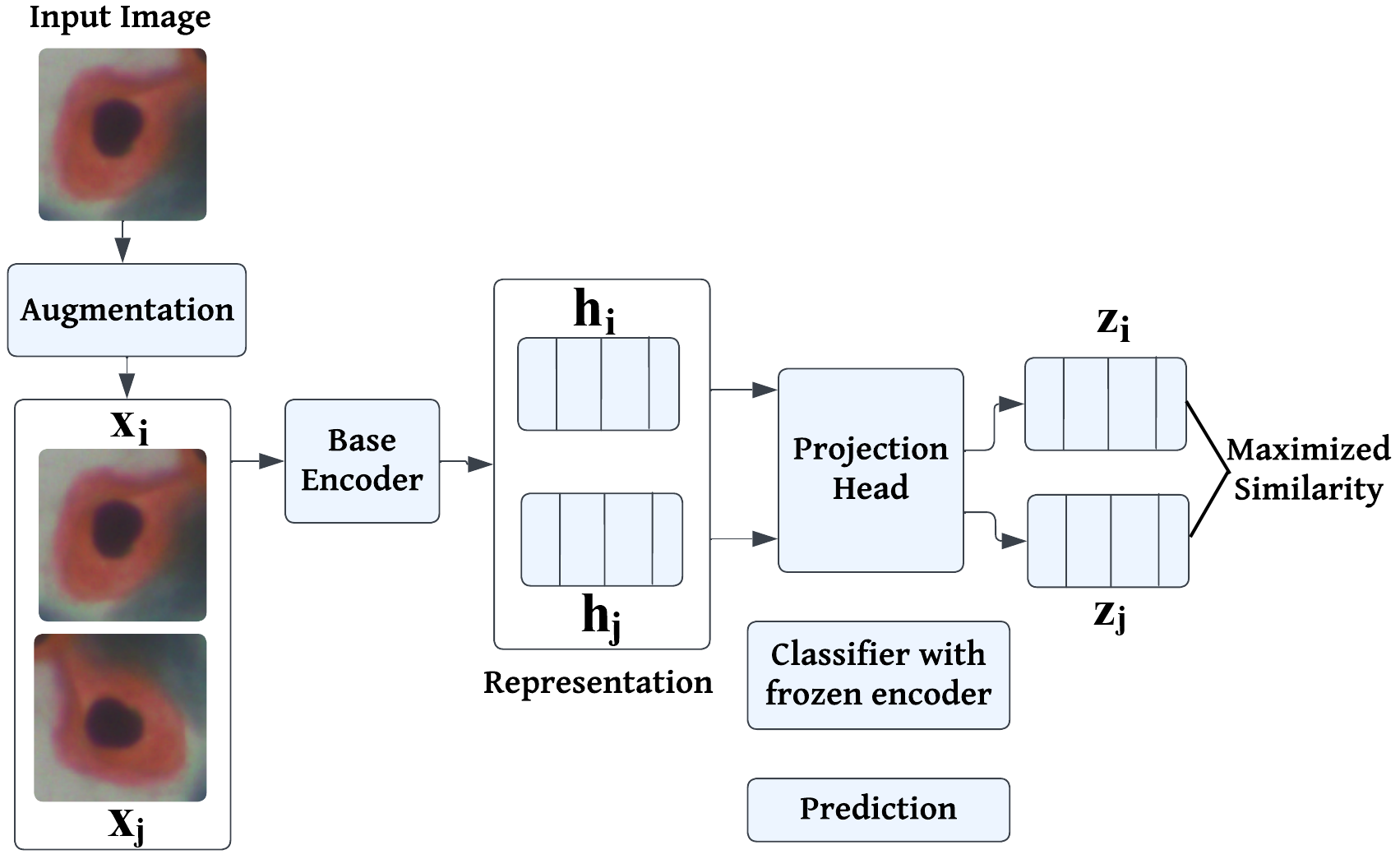}
    \caption{Steps of Supervised Contrastive Learning}
    \label{fig:contrastive}
\end{figure}

Supervised contrastive learning\cite{b9} focuses on learning the representation of the instances in the dataset by minimizing similarity between negative pairs and maximizing similarity between positive pairs in the feature space. The contrastive loss function is designed to pull together similar instances in the feature space while pushing apart dissimilar instances. The steps are depicted in Figure \ref{fig:contrastive}.
At first, we augment the input images. Images with the same class labels are considered positive pairs, and images with different class labels are considered negative pairs. The images of each batch are then passed through a base encoder, which is a pre-trained CNN model. The base encoder extracts high-level features from the input images and maps them into a continuous representation space. The representations obtained from the base encoder are then fed into a projection head. The projection head consists of fully connected layers that project the representations into a different space. This projection step helps to maximize the similarity between the representations of positive pairs while minimizing the similarity between representations of negative pairs. After the projection, the representations are passed through a frozen encoder. The frozen encoder is a copy of the base encoder with its weights fixed. It serves as a feature extractor, aiming to capture additional information from the representations. Finally, the frozen encoder's output is used to predict the labeled data.

\subsection{Performance Evaluation}
To evaluate our system, we have used accuracy, weighted precision, weighted recall, and weighted F1 score. The weighted metrics were measured by considering the instance occurrences of the classes in the dataset.

\begin{table}[ht]
\caption{Performance Comparison using Fine-Tuning}
\label{tab:finetuning}
\centering
\resizebox{\columnwidth}{!}
{\begin{tabular}{|p{1.22cm}|c|c|c|c|p{0.9cm}|}
\hline
\textbf{Classifier} & \textbf{Class} & \textbf{Weighted} & \textbf{Weighted} & \textbf{Weighted} & \textbf{Accuracy} \\
 &  & \textbf{Precision} & \textbf{Recall} & \textbf{F1 Score} & \\
\hline\rule{0pt}{12pt}
& Dyskeratotic & 0.94 & 0.96 & 0.95& \\
& Koilocytotic & 0.93 & 0.88 & 0.90&\\
VGG16 & Metaplastic & 0.95 & 0.96 & 0.96 & 95.57\% \\
& Parabasal & 0.99 & 0.99 & 0.99&\\
& Superficial- & 0.97 & 0.99 & 0.98 &\\
& Intermediate & & & & \\
\hline\rule{0pt}{12pt}
& Dyskeratotic & 0.92 & 0.94 & 0.93& \\
& Koilocytotic & 0.90 & 0.87 & 0.88&\\
VGG19 & Metaplastic & 0.94 & 0.94 & 0.94 & 94.21\% \\
& Parabasal & 0.97 & 0.98 & 0.98&\\
& Superficial- & 0.98 & 0.99 & 0.99 &\\
& Intermediate & & & & \\
\hline\rule{0pt}{12pt}
& \textbf{Dyskeratotic} & \textbf{0.95} & \textbf{0.97} & \textbf{0.96} & \\
& \textbf{Koilocytotic} & \textbf{0.98} & \textbf{0.93} & \textbf{0.96} & \\
\textbf{DenseNet-169} & \textbf{Metaplastic} & \textbf{0.95} & \textbf{0.98} & \textbf{0.97} & \textbf{97.17\%} \\
& \textbf{Parabasal} & \textbf{0.99} & \textbf{0.99} & \textbf{0.99} & \\
& \textbf{Superficial-} & \textbf{0.99} & \textbf{0.99} & \textbf{0.99} &\\
& \textbf{Intermediate} & & & & \\
\hline\rule{0pt}{12pt}
& Dyskeratotic & 0.98 & 0.96 & 0.97& \\
& Koilocytotic & 0.93 & 0.94 & 0.93&\\
MobileNet-V2 & Metaplastic & 0.98 & 0.95 & 0.97 & 96.55\% \\
& Parabasal & 0.97 & 1.00 & 0.98&\\
& Superficial- & 0.98 & 0.98 & 0.98 &\\
& Intermediate & & & & \\
\hline\rule{0pt}{12pt}
& Dyskeratotic & 0.94 & 0.96 & 0.95& \\
& Koilocytotic & 0.89 & 0.90 & 0.89&\\
ResNet-50 & Metaplastic & 0.92 & 0.93 & 0.93 & 94.58\% \\
& Parabasal & 1.00 & 0.98 & 0.99 &\\
& Superficial- & 0.99 & 0.96 & 0.98&\\
& Intermediate & & & & \\
\hline
\end{tabular}}
\end{table}

\subsection{Interpretation} In the final step, we leverage gradient-based and perturbation-based explainable AI techniques to make our system transparent and reliable. Gradient-based Explainability Analysis (XAI) uses gradients to understand how input features contribute to model predictions. GradCAM\cite{b23} is a class-discriminative localization technique that computes the importance score in the final convolutional layer of a CNN using gradients. GradCAM++\cite{b24} takes both forward and backward gradients into account, whereas ScoreCAM\cite{b25} employs a global average pooling operation to calculate a single importance score for each feature map. LayerCAM\cite{b26} generates class activation maps by analyzing different layers of a deep neural network and identifying regions in an input image that contribute significantly to the model's classification decision. Perturbation-based XAI methods involve altering input data to observe the model's predictions. LIME, or Local Interpretable Model Agnostic Explanation, generates random perturbations to determine feature significance in classification\cite{b27}. It is model-agnostic, highlights positive superpixels, and uses color-coded visualizations to provide interpretable insights into the model's decision-making process.

\section{Evaluation}
\subsection{Hyper-Parameter Setting}
All the experiments were performed on Google Colaboratory and Kaggle Notebook. The implementations were written using Python language, Tensorflow framework, and Keras Library. Conducting a thorough experimentation and hyperparameter tuning, we obtained optimal results with \textit{Softmax} activation, \textit{Adam} optimizer with a learning rate of $1\times10^{-3}$, and a dropout rate of 0.50. Conducting all experiments for 50 epochs resulted in consistently better performance, establishing model convergence. While we incorporated \textit{Categorical cross-entropy} for standard fine-tuning and fine-tuning with cost-sensitive learning, we utilized multi-class \textit{n-pairs loss} for fine-tuning with contrastive learning. We also experimented with supervised \textit{NT-Xent loss}, \textit{triplet margin loss}, but multi-class \textit{n-pairs loss} yields the best performance.


\subsection{Experimental Results}
We conducted three training experiments: standard fine-tuning, fine-tuning with cost-sensitive learning, and fine-tuning with supervised contrastive learning. The results are summarized in Tables \ref{tab:finetuning},\ref{tab:tab2}, and \ref{tab:tab3}.
In the initial fine-tuning phase, as indicated in Table \ref{tab:finetuning}, DenseNet-169 emerged as the top-performing model with an impressive accuracy of 97.17\%. Conversely, VGG19 and ResNet-50 exhibited relatively lower performance, while VGG16 and MobileNetV2 achieved intermediate results.
Moving on to fine-tuning with cost-sensitive learning, as shown in \ref{tab:tab2}, DenseNet-169 maintained its superiority but with only a slight improvement in its performance. However, in the case of fine-tuning with supervised contrastive learning, as shown in \ref{tab:tab3}, DenseNet-169's performance took a downturn, with VGG-16 emerging as the best classifier with a remarkable accuracy of 97.29\%.

\begin{table}[ht]
\caption{Performance Comparison using Cost-Sensitive Learning}
\label{tab:tab2}
\centering
\resizebox{\columnwidth}{!}
{\begin{tabular}{|p{1.22cm}|c|c|c|c|p{0.9cm}|}
\hline
\textbf{Classifier} & \textbf{Class} & \textbf{Weighted} & \textbf{Weighted} & \textbf{Weighted} & \textbf{Accuracy} \\
 &  & \textbf{Precision} & \textbf{Recall} & \textbf{F1 Score} & \\
\hline\rule{0pt}{12pt}
& Dyskeratotic & 0.96 & 0.98 & 0.97& \\
& Koilocytotic & 0.94 & 0.90 & 0.92&\\
VGG16 & Metaplastic & 0.95 & 0.95 & 0.95 & 95.94\% \\
& Parabasal & 0.99 & 0.99 & 0.99&\\
& Superficial- & 0.97 & 0.98 & 0.97&\\
& Intermediate & & & & \\
\hline\rule{0pt}{12pt}
& Dyskeratotic & 0.96 & 0.93 & 0.94& \\
& Koilocytotic & 0.88 & 0.90 & 0.89&\\
VGG19 & Metaplastic & 0.93 & 0.91 & 0.92 & 94.09\% \\
& Parabasal & 0.97 & 0.98 & 0.98&\\
& Superficial- & 0.97 & 0.99 & 0.98&\\
& Intermediate & & & & \\
\hline\rule{0pt}{12pt}
 & \textbf{Dyskeratotic} & \textbf{0.96} & \textbf{0.96} & \textbf{0.96} & \\
 & \textbf{Koilocytotic} & \textbf{0.96} & \textbf{0.94} & \textbf{0.95} & \\
\textbf{DenseNet-169} & \textbf{Metaplastic} & \textbf{0.96} & \textbf{0.98} & \textbf{0.97} & \textbf{97.29\%} \\
 & \textbf{Parabasal} & \textbf{1.00} & \textbf{0.99} & \textbf{0.99} & \\
 & \textbf{Superficial-} & \textbf{0.99} & \textbf{0.99} & \textbf{0.99} & \\
 & \textbf{Intermediate} & & & & \\
\hline\rule{0pt}{12pt}
& Dyskeratotic & 0.98 & 0.94 & 0.96& \\
& Koilocytotic & 0.90 & 0.94 & 0.92&\\
MobileNet-V2 & Metaplastic & 0.97 & 0.93 & 0.95 & 95.57\% \\
& Parabasal & 0.97 & 1.00 & 0.98&\\
& Superficial- & 0.96 & 0.97 & 0.97 &\\
& Intermediate & & & & \\
\hline\rule{0pt}{12pt}
& Dyskeratotic & 0.91 & 0.98 & 0.94& \\
& Koilocytotic & 0.91 & 0.87 & 0.89&\\
ResNet-50 & Metaplastic & 0.92 & 0.92 & 0.92 & 94.21\% \\
& Parabasal & 0.99 & 0.97 & 0.98 &\\
& Superficial- & 0.98 & 0.97 & 0.97&\\
& Intermediate & & & & \\
\hline
\end{tabular}}
\end{table}



\begin{table}[ht]
\caption{Performance Comparison using Contrastive Learning}
\label{tab:tab3}
\centering
\resizebox{\columnwidth}{!}
{\begin{tabular}{|p{1.22cm}|c|c|c|c|p{0.9cm}|}
\hline
\textbf{Classifier} & \textbf{Class} & \textbf{Weighted} & \textbf{Weighted} & \textbf{Weighted} & \textbf{Accuracy} \\
 &  & \textbf{Precision} & \textbf{Recall} & \textbf{F1 Score} & \\
\hline\rule{0pt}{12pt}
 & \textbf{Dyskeratotic} & \textbf{0.95} & \textbf{0.98} & \textbf{0.97} & \\
 & \textbf{Koilocytotic} & \textbf{0.98} & \textbf{0.94} & \textbf{0.96} & \\
\textbf{VGG16} & \textbf{Metaplastic} & \textbf{0.96} & \textbf{0.97} & \textbf{0.97} & \textbf{97.29\%} \\
 & \textbf{Parabasal} & \textbf{1.00} & \textbf{0.99} & \textbf{0.99} & \\
 & \textbf{Superficial-} & \textbf{0.98} & \textbf{0.99} & \textbf{0.98} & \\
 & \textbf{Intermediate} & & & & \\
\hline\rule{0pt}{12pt}
& Dyskeratotic & 0.98 & 0.97 & 0.97& \\
& Koilocytotic & 0.95 & 0.95 & 0.95&\\
VGG19 & Metaplastic & 0.94 & 0.96 & 0.95 & 96.68\% \\
& Parabasal & 0.99 & 0.97 & 0.98&\\
& Superficial- & 0.99 & 0.98 & 0.99&\\
& Intermediate & & & & \\
\hline\rule{0pt}{12pt}
& Dyskeratotic & 0.96 & 0.95 & 0.95& \\
& Koilocytotic & 0.90 & 0.86 & 0.88&\\
DenseNet-169 & Metaplastic & 0.92 & 0.93 & 0.92 & 93.47\% \\
& Parabasal & 0.93 & 0.98 & 0.95&\\
& Superficial- & 0.96 & 0.96 & 0.96&\\
& Intermediate & & & & \\
\hline\rule{0pt}{12pt}
& Dyskeratotic & 0.95 & 0.96 & 0.95 & \\
& Koilocytotic & 0.95 & 0.93 & 0.94 & \\
MobileNet-V2 & Metaplastic & 0.95 & 0.97 & 0.96 & 95.81\% \\
& Parabasal & 0.97 & 0.99 & 0.98&\\
& Superficial- & 0.98 & 0.96 & 0.97&\\
& Intermediate & & & & \\
\hline\rule{0pt}{12pt}
& Dyskeratotic & 0.95 & 0.96 & 0.95 & \\
& Koilocytotic & 0.95 & 0.93 & 0.94 & \\
ResNet-50 & Metaplastic & 0.96 & 0.98 & 0.97 & 96.43\% \\
& Parabasal & 0.98 & 0.99 & 0.99 & \\
& Superficial- & 0.98 & 0.96 & 0.97 & \\
& Intermediate & & & & \\
\hline
\end{tabular}}
\end{table}

\begin{table}[ht]
\caption{Comparison with previous works}
\label{tab:comparison}
\centering
\begin{tabular}{|p{3cm}|p{3cm}|c|}
\hline
\textbf{Author} & \textbf{Methods} & \textbf{Accuracy}\\
\hline
Pramanik et al. \cite{b5} & Fuzzy distance-based ensemble approach using Inception V3, Inception ResNet V2, and MobileNet V2 & 96.96\%\\
\hline
A. Manna, R. Kundu, D. Kaplun, A. Sinitca, and R. Sarkar \cite{manna2021fuzzy}  & Ensemble technique using Inception v3, DenseNet-169, and Xception & 95.43\%\\
& & \\
\hline
A. Tripathi, A. Arora, and A. Bhan \cite{b4} & Fine-tuning (ResNet-152) & 94.89\%\\
\hline
\textbf{Our study} & Fine-tuning (DenseNet-169) & 97.17\%\\
 & \textbf{Cost-Sensitive learning (DenseNet-169)} & \textbf{97.29}\%\\
 & \textbf{Supervised Contrastive learning (VGG16)} & \textbf{97.29}\%\\
\hline
\end{tabular}
\end{table}

\subsection{Comparison with Existing Works}
Comparing our study to existing works (Table \ref{tab:comparison}), it is evident that our approach outperforms several prior studies. Pramanik et al. achieved 96.96\% accuracy using an ensemble approach, A. Manna, R. Kundu, D. Kaplun, A. Sinitca, and R. Sarkar reached 95.43\% with another ensemble technique, and A. Tripathi, A. Arora, and A. Bhan obtained 94.89\% through ResNet-152 fine-tuning. Our study surpassed these benchmarks, achieving 97.17\% accuracy with DenseNet-169 fine-tuning. DenseNet-169 with cost-sensitive learning further improved the accuracy to 97.29\%, while in contrastive learning, VGG16 excelled with an accuracy of 97.29\% as well, showcasing superior performance compared to prior work.

\subsection{Interpretation using XAI}
We used neural network visualization toolkit\footnote{\url{https://github.com/keisen/tf-keras-vis}} to use gradient-based visualization techniques like GradCAM, GradCAM++, ScoreCAM, LayerCAM, and perturbation-based visualization technique LIME to visualize classifications.
\begin{figure}[htbp]
    \centering
    \includegraphics[width=85mm,scale=1]{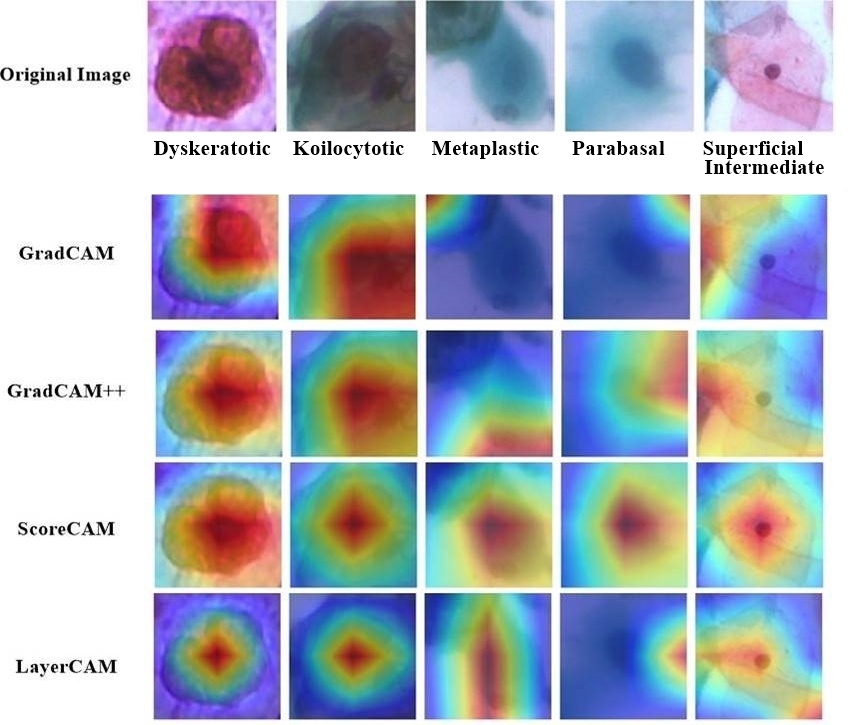}
    \caption{Five sample outputs of correctly classified instances of DenseNet-169 using gradient-based XAI techniques}
    \label{fig:FT_Cor}
\end{figure}

\begin{figure}[htbp]
    \centering
    \includegraphics[width=85mm,scale=1]{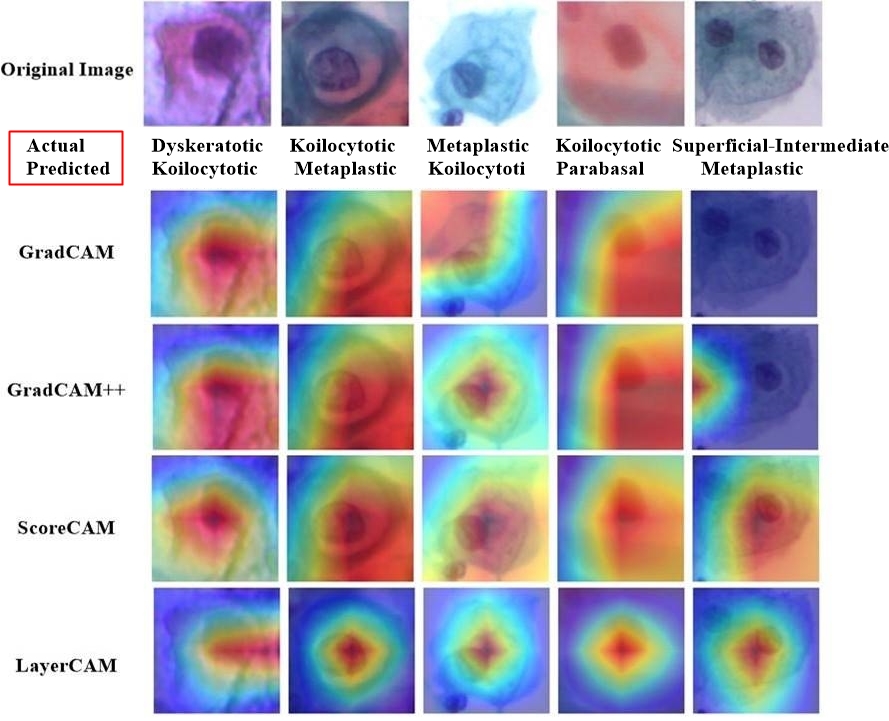}
    \caption{Five sample outputs of misclassified instances of DenseNet-169 using gradient-based XAI techniques}
    \label{fig:FT_Mis}
\end{figure}
Figure \ref{fig:FT_Cor} represents five correctly classified images from five different classes, and \ref{fig:FT_Mis} represents five misclassified images where advanced gradient-based visualization techniques were employed, including GradCAM, GradCAM++, ScoreCAM, and LayerCAM. By applying these methods, crucial areas within the images were highlighted, shedding light on the regions that influenced the classification. In \ref{fig:FT_Cor}, GradCAM and GradCAM++ were unable to generate satisfactory heatmaps for the \textit{Metaplastic} and \textit{Parabasal} classes. LayerCAM also struggled to produce an accurate heatmap for the \textit{Parabasal} class. However, ScoreCAM proved to be effective in generating a reliable heatmap for the \textit{Parabasal} class. The limitations observed in the performance of GradCAM, GradCAM++, and LayerCAM emphasize their difficulty in capturing the essential features and influential regions within the \textit{Metaplastic} and \textit{Parabasal} class images. By using a combination of techniques like GradCAM, GradCAM++, LayerCAM, and ScoreCAM, we were able to gain a comprehensive understanding of the model's behavior across different classes, identifying both strengths and weaknesses in its classification capabilities.
\subsection{Perturbation-based Visualization}
\begin{figure}[htbp]
    \centering
    \includegraphics[width=85mm,scale=1]{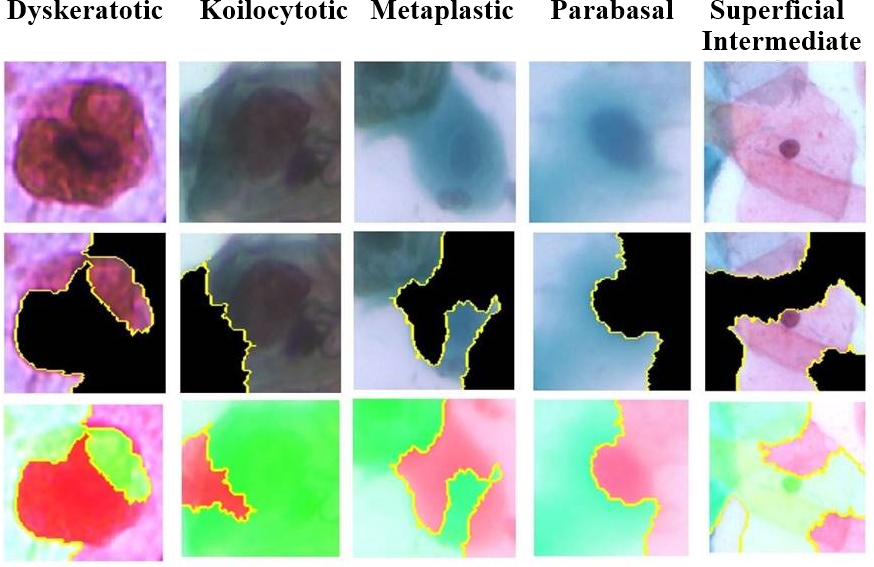}
    \caption{Five sample outputs of correctly classified instances of DenseNet-169 using Perturbation-based Visualization technique LIME}
    \label{fig:FT_Cor_LIME}
\end{figure}
\begin{figure}[htbp]
    \centering
    \includegraphics[width=85mm,scale=1]{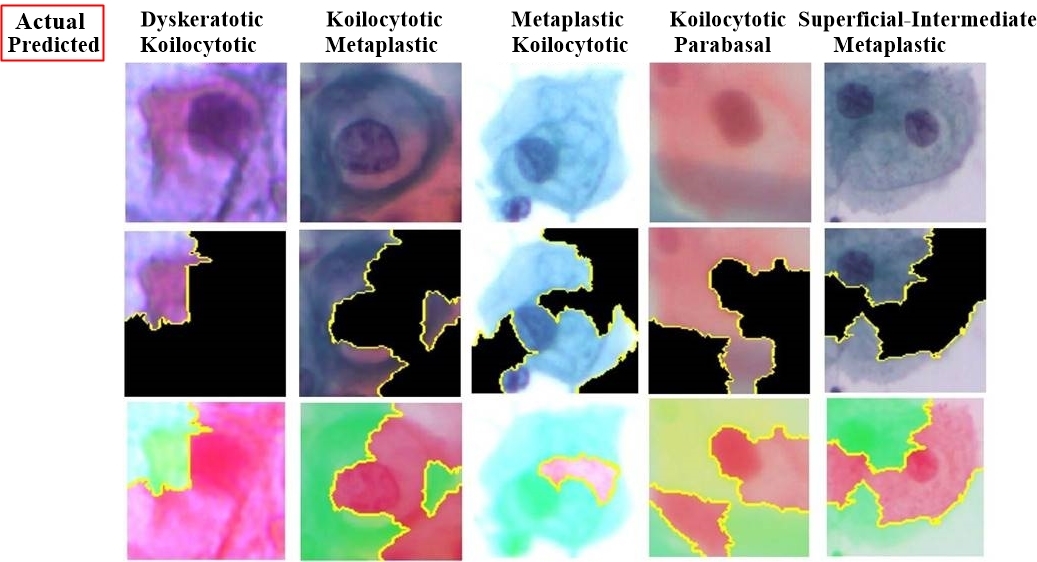}
    \caption{Five sample outputs of misclassified instances of DenseNet-169 using Perturbation-based Visualization technique LIME}
    \label{fig:FT_Mis_LIME}
\end{figure}
Figure \ref{fig:FT_Cor_LIME} displays a set of five correctly classified images representing five different classes, while Figure \ref{fig:FT_Mis_LIME} presents five misclassified images where the perturbation-based visualization technique, Local Interpretable Model-Agnostic Explanations (LIME) was used. LIME helped identify and highlight the important regions and features within these images that contributed to the incorrect classifications. By generating random perturbations of the input data, LIME identified the top superpixels that positively influence the predicted class. LIME generated two types of visualizations of the images: one that focuses solely on positive contributions and another that incorporates both positive and negative contributions. These visualizations allowed us to gain insights into potential weaknesses or biases present in the model's decision-making process. By highlighting the influential regions while preserving the rest of the image, LIME provides interpretable insights into the factors influencing the model's output. Moreover, LIME employed color-coded visualizations, with green representing the "pros" (positive contributions) and red representing the "cons" (negative contributions). This color scheme facilitates a clear understanding of the model's decision by visualizing the positive and negative aspects of the prediction. LIME's approach also gives interpretable insights into the model's decision-making process by showcasing the pixels that influenced the misclassifications.

\section{Conclusion}
This study utilizes deep learning techniques, including VGG16, VGG19, ResNet-50, MobileNetV2, and DenseNet-169, with a focus on cost sensitivity and supervised contrastive learning to improve cervical cancer cell classification. Evaluation metrics like precision, recall, F1 score, and accuracy are used to assess the system performance. We enhance interpretability with gradient and perturbation-based visualization, aiding trust in automated decisions. Our research showcases the potential of automated systems in cervical cancer detection, contributing to early prevention. Future work could involve broader dataset testing for generalization performance.

\bibliographystyle{IEEEtran}
\bibliography{reference}
\end{document}